%% file: paper.tex
\documentclass{article}
\pdfoutput=1
\usepackage[preprint]{spconf}
\usepackage{adjustbox, amsmath, amssymb, graphicx, multirow, booktabs}
\usepackage{algorithm, algpseudocode}
\usepackage{url}
\makeatletter
\providecommand\add@text{}
\newcommand\tagaddtext[1]{%
  \gdef\add@text{#1\gdef\add@text{}}}%
\renewcommand\tagform@[1]{%
  \maketag@@@{\llap{\add@text\quad}(\ignorespaces#1\unskip\@@italiccorr)}%
}
\makeatother

\usepackage{pgfplots}
\pgfplotsset{grid style={dashed,gray}}
\pgfplotsset{compat=1.12}
\usepackage{tikz}
\usetikzlibrary{shapes.geometric}
\usetikzlibrary{shapes.misc,shadows}
\usetikzlibrary{positioning} 
\usetikzlibrary{arrows} 
\usetikzlibrary{arrows.meta}
\tikzset{%
    >={Latex[width=1mm,length=1mm]},
    base/.style = {
        rectangle, rounded corners, draw=black,
        minimum width=2cm, minimum height=.4cm,
        text centered, font=\tiny},
    acoustic_model/.style = {base, fill=red!15},
    language_model/.style = {base, fill=cyan!20},
    joint/.style = {base, fill=yellow!15},
    io/.style = {base, fill=none, draw=none, minimum width=0cm},
    data/.style = {
        rectangle, draw, align=center, left color=blue!20, right color=white,
        minimum width=0.5cm, minimum height=0.5cm},
    block/.style ={
        rectangle, thick, draw=black, align=center, fill=orange!15,
        minimum height=.3cm, minimum width=2cm, text width=2cm},
    connector/.style={-latex, font=\tiny},
    rectangle connector/.style={
        connector,
        to path={(\tikztostart) -- ++(#1,0pt) \tikztonodes |- (\tikztotarget) },
        pos=0.5
    },
}

\usepackage{color}

\copyrightnotice{\copyright\ IEEE 2000}

\title{Training Speech Recognition Models with Federated Learning: \\A Quality/Cost Framework}

\name{Dhruv Guliani \qquad Fran\c{c}oise Beaufays \qquad Giovanni Motta}
\address{Google, Mountain View, CA, U.S.A \\
{dguliani@google.com}}

\begin{document}
%
\maketitle
\begin{abstract}
We propose using federated learning, a decentralized on-device learning paradigm, to train speech recognition models. By performing epochs of training on a per-user basis, federated learning must incur the cost of dealing with non-IID data distributions, which are expected to negatively affect the quality of the trained model. We propose a framework by which the degree of non-IID-ness can be varied, consequently illustrating a trade-off between model quality and the computational cost of federated training, which we capture through a novel metric. Finally, we demonstrate that hyper-parameter optimization and appropriate use of variational noise are sufficient to compensate for the quality impact of non-IID distributions, while decreasing the cost.
\end{abstract}
\begin{keywords}
speech recognition, federated learning.
\end{keywords}
\input{introduction}
\input{methodology}
\input{model}

\input{experiments}
\input{conclusions}
\input{acknowledgements}

\vfill\pagebreak

\bibliographystyle{IEEEbib}
\bibliography{refs}

\end{document}

%% file: introduction.tex
\section{Introduction}
\label{sec:intro}

The increasing ubiquity of mobile devices with high computational power~\cite{device_ownership}, along with advances in sequence-to-sequence neural networks~\cite{st_rnn, sequence_models_in_asr}, have made it possible to develop mobile applications powered by on-device automatic speech recognition (ASR) systems~\cite{e2e_surpasses_server}. For example, neural ASR models with state-of-the-art quality~\cite{e2e_streaming} have been deployed on-device with additional latency and reliability benefits~\cite{e2e_asr_blogpost} relative to server-based models.

Considering user privacy in the context of on-device ASR, we investigate the feasibility of {\em{training}} speech models on-device using federated learning (FL)~\cite{fedavg, fl_at_scale}. FL is a decentralized training method that does not require sending raw user data to central servers. Instead, user data are stored in an on-device {\em{training cache}}, where training iterations can be performed. FL optimization proceeds in synchronous \emph{rounds} of training, wherein a set of \emph{clients} (devices) contributes updates to a central model. FL has been successfully deployed in large production systems to perform emoji prediction~\cite{fedemoji}, next-word prediction~\cite{fed_nextword}, and query suggestion~\cite{fed_query}.

When used to perform per-user training of models, FL presents an inherent difference in comparison to centralized training in that training data under FL are non-IID. With standard central mini-batch training, an IID (independent and identically distributed) assumption is typically made as examples are sampled with a uniform probability across the training set. Under FL, examples are sampled from client distributions which may be similar, but not identical (non-IID). Training on non-IID data has been shown to be sub-optimal across multiple domains~\cite{noniid_fedvision, noniid_keyword, noniid_divergence} and remains an open problem in federated learning~\cite{fl_open_problems, noniid_quagmire}.

In this work, we study the impact of speaker-split (non-IID) data in the context of ASR training, and make the following contributions:
\begin{itemize}
    \itemsep0em
    \item We provide a mental model to reason about the differences between IID and non-IID training.
    \item We introduce a general \emph{cost} function for FL which measures the computational cost of model quality.
    \item We show that ASR models trained with FL can achieve the same quality as server-trained models.
\end{itemize}

%% file: methodology.tex
\section{Methodology}
\label{sec:methodology}

\subsection{Federated Averaging Algorithm}
\label{ssec:fedavg}

A common optimization algorithm for FL is Federated Averaging (\emph{FedAvg})~\cite{fedavg}. Outlined in Algorithm 1,~\emph{FedAvg} consists of two levels of optimization: local optimization performed on $K$ participating clients, and a server step to update the global model. \emph{FedAvg} is used in all experiments in this work.

\begin{algorithm}
\caption{\emph{FedAvg}. The $K$ clients are indexed by $k$, rounds are indexed by $r$, and $n$ is the number of examples.}
\begin{algorithmic}[1]\label{alg:fedavg}
\State initialize $w^0$
\For{each round $r$ = 1,2,...}
    \State{$K \gets$ (random subset of $M$ clients)}
    \For{each client $k \in K$ \textbf{in parallel}}
        \State{$\hat{w}^r_k \gets Client Update(k, w^r)$}
        \State{$\Delta w^r_k = w^r - \hat{w}^r_k$}
    \EndFor
    \State{$\bar{w}^r = \sum_{k=1}^{K} \frac{n_k}{n} \Delta w^r_k$} \Comment{Weighted average}
    \State{$w^{r+1} = w^r - \eta\bar{w}^r$  \Comment{Server update}}
\EndFor
\end{algorithmic}
\end{algorithm}

\vspace{-0.2in}
\subsection{Understanding non-IID Data}
\label{ssec:fedavg_noniid}

Various studies have shown noticeable quality degradation when training neural models with non-IID data. Strategies to mitigate this include accounting for client model drift~\cite{noniid_divergence}, using adaptive optimizers~\cite{adaptive_optimizers}, tuning local optimization hyper-parameters~\cite{noniid_fedvision, fedavg}, and weighting client updates with estimates of client data skew~\cite{noniid_quagmire, fed_dga, fedvision_mitigating_noniid}.

In this work, we build on the observation that, given an increased computation budget, a non-IID distribution can be modified to approximate an IID one. To elaborate, in a federated training round, a set of clients is randomly selected from a training population. If, in the client optimization, a single local example is sampled and used to compute an ~\emph{SGD} update with learning rate 1, raw gradients would be aggregated in the server update step. As the sampling of clients in a federated training round is random, the server step would aggregate effectively IID contributions in this example. However, training in this manner would likely cause a sharp increase in the number of rounds needed for convergence, and subsequently increase training time and server-client communication.

We therefore note that the degree of non-IID-ness in a particular experiment can be varied at the expense of computational cost, and this idea can be used to make adjustments to the experimental setup depending on the desired quality-performance trade-off.

\subsection{The Cost of Federated Model Quality}
\label{ssec:cost_func}

We capture the cost of federated computation through a metric we name the \emph{Cost of Federated Model Quality} (\textrm{CFMQ}). This is, to our knowledge, the first attempt at formulating a general cost function which may be adapted for any federated optimization. Along with non-IID/IID considerations, this cost function helps compare the impact of convergence time, local optimization, client participation and communication payload on quality. Therefore, when used in conjunction with a quality metric, the \textrm{CFMQ} provides a useful way to compare experiments.

Let $\mu$ be the average number of local optimization steps taken by a client. Let $e$ be the number of local epochs, $N$ the total number of examples in a round, $b$ the batch size, and $K$ the number of clients participating. It follows that:
\begin{equation}
\label{eqn:num_client_steps}
\mu = \frac{eN}{bK}
\end{equation}

Let $P$ be the round-trip communication payload in bytes, $R$ the number of rounds, and $\nu$ the peak memory consumed during a step. Equation~\ref{eqn:fl_cost} unifies the communication cost and local computation cost, as these are the two resource-constrained aspects of the federated optimization~\cite{fed_query}. We assume an abundance of server resources/memory in our study. We therefore formulate the cost \textrm{CFMQ} as:

\begin{equation}
\label{eqn:fl_cost}
\textrm{CFMQ} = RK (P + \alpha\mu \nu)
\tagaddtext{[{bytes}]}
\end{equation}

\noindent $\alpha$ is a balancing term added to the \textrm{CFMQ}, and can be modified to adjust the importance of the two components of cost.

%% file: model.tex
\section{Model and Data}
\label{sec:model}

\subsection{Model Architecture}
\label{ssec:rnnt}

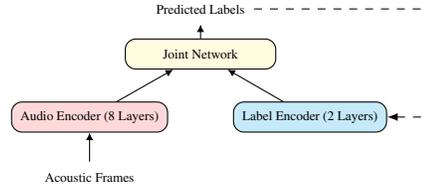
\begin{figure}[t]
\centering
\begin{tikzpicture}[node distance=.6cm, every node/.style={fill=white}, align=center]
  \node (output)             [io]              {Predicted Labels};
  \node (joint)     [joint, below of=output]          {Joint Network};

  \node (am)     [acoustic_model, below left=of joint, xshift=1cm]      {Audio Encoder (8 Layers)};
  \node (lm)     [language_model, below right=of joint, xshift=-1cm]    {Label Encoder (2 Layers)};
  \node (input)  [io, below of=am, yshift=-0.2cm]           {Acoustic Frames};
  \draw[->]             (input) -- (am);
  \draw[->]             (am) -- (joint);
  \draw[->]             (lm) -- (joint);
  \draw[->]             (joint) -- (output);
  \draw [rectangle connector=3cm, dashed] (output) to (lm);

\end{tikzpicture}
\caption{RNN-T speech model architecture.}
\label{fig:rnnt_summary}
\end{figure}

We use the RNN-T architecture~\cite{e2e_streaming} depicted in Figure~\ref{fig:rnnt_summary} in this paper. The model has 122M trainable parameters, and predicts the probability $P(y|x)$ of labels $y$ given acoustic data $x$. It consists of an LSTM audio encoder, an LSTM label encoder, a fully-connected layer concatenating the encoder outputs, and an output softmax.  The input acoustic frames are 128-dimensional log-mel filterbank energies, and output labels belong to a set of 4096 word-pieces.

\subsection{Librispeech Corpus}
\label{ssec:libri}

We use the Librispeech~\cite{libri} corpus, containing ~960 hours of transcribed training utterances from 2338 speakers, and ~21 hours of evaluation audio from 146 speakers split amongst 4 sets \emph{Dev}, \emph{DevOther}, \emph{Test}, and \emph{TestOther}, with the reporting metric of~\emph{Word Error Rate} (WER). The sets labelled ``Other'' are intended to be more difficult to recognize. The data are evenly balanced in terms of male and female speakers.

 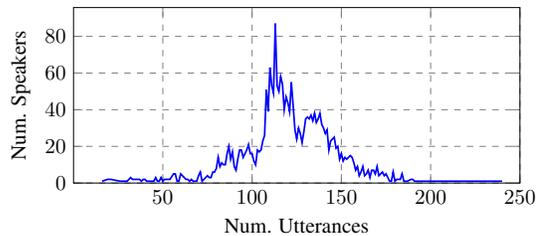
\begin{figure}[ht]
    \centering
    \begin{tikzpicture}[scale=.8]
    \begin{axis}[
        height=4.5cm,
		width=9cm,
        legend pos=north east,
        xmin=0,
        xmax=250,
        ymin=0,
        grid=both,
        xtick={50,100,150,200,250},
    	ytick={0,20,40,60,80},
    	xlabel={Num. Utterances},
    	ylabel={Num. Speakers}
    	]
	\addplot[thick,blue]
        table[x=num_utterances,y=num_speakers,col sep=comma]{libri_speaker_dist.csv};
    \end{axis}
    \end{tikzpicture}
    \vspace{-0.1in}
    \caption{Histogram of utterance distribution across speakers.}
    \label{fig:libri_hist}
\end{figure}

We run FL experiments on this corpus by associating each speaker label to a device that could participate in rounds of federated training. Librispeech data, when split by speaker, are non-IID for a variety of reasons including differences in voice, vocabulary, recording quality, and utterance counts (as represented in Figure~\ref{fig:libri_hist}) across users.

%% file: experiments.tex
\section{Experiments}
\label{sec:experiments}

\subsection{Baseline and Federated Training}
\label{ssec:baseline}

We conducted a series of experiments to recover quality degradation due to non-IID training, and compared them to a centrally trained IID Baseline. The Baseline configuration was trained with a linear ramp-up learning rate schedule, SpecAugment~\cite{specaug}, and Variational Noise~\cite{variational_inference}. Baseline results (\emph{E0}) can be found in Table~\ref{tab:non_iid_first}.

Federated training of RNN-T models was performed using \emph{FedAvg} on a FL simulator written in TensorFlow~\cite{tensorflow} running on TPU~\cite{tpu} hardware, where data were split by speaker. Training hyper-parameters were kept as similar to the Baseline as possible, with the exception of Variational Noise initially being omitted as it needed to be adapted for the FL (discussed further in~\ref{sssec:federated_vn}).

\subsection{Matching Baseline Model Quality}
\label{ssec:match_baseline}

In non-IID federated experiments,~\emph{SGD} was used as the client optimizer with a constant learning rate which was set to $0.008$ through a coarse sweep. Adam~\cite{adaptive_optimizers} was used for the server update. The number of participating clients, $K$, was gradually increased from 32 to 128, beyond which it stopped offering improvements to model quality. In this configuration, clients cycled through local data over a single epoch.
\begin{table}[htpb]
    \centering
    \begin{adjustbox}{center, width=\columnwidth-20pt}
    \begin{tabular}{c | c | c c c c}
    \toprule
        \multirow{2}*{\textbf{ID}}
        & \multirow{2}*{\textbf{Exp.}}
        & \multicolumn{4}{c}{\textbf{WER}} \\
        & & \textit{Test} & \textit{TestOther} & \textit{Dev} & \textit{DevOther} \\
    \midrule
        {E0} & {Baseline}               & 4.8 & 12.1 & 5.1 & 12.1 \\
    \midrule
        {E1} & {Non-IID}                & 6.8 & 17.2 & 7.0 & 17.3 \\
    \midrule
        \multicolumn{2}{c|}{\textit{E0 v. E1 \% Rel. WER}}
            & {+42\%} & {+42\%} & {+37\%} & {+43\%} \\
    \bottomrule
    \end{tabular}
    \end{adjustbox}
    \caption{Quality degradation with non-IID training.\label{tab:non_iid_first}}
\end{table}
Table~\ref{tab:non_iid_first} shows the Baseline performance and the initial non-IID config, with a substantial WER degradation across all evaluation sets.

\subsubsection{Limiting Per-speaker Data}
\label{sssec:speaker_data_limit}

In an attempt to push data distributions closer to IID and minimize per-client drift, a subset of examples were randomly sampled from each speaker participating in a federated round to impose a per-client data limit. We show that data-limiting pushes distributions to be more IID through a thought experiment, wherein a single example is sampled from each client. In this scenario, assuming client-selection is random, the participating data in a round is as close as possible to IID.

It is important to note that the entire per-speaker dataset was still seen over the course of multiple rounds.

\begin{table}[htpb]
    \centering
    \begin{adjustbox}{center, width=\columnwidth-20pt}
    \begin{tabular}{c | c | c c c c}
    \toprule
        \multirow{2}*{\textbf{ID}}
        & \multirow{2}*{\textbf{Data Limit}}
        & \multicolumn{4}{c}{\textbf{WER}} \\
        & & \textit{Test} & \textit{TestOther} & \textit{Dev} & \textit{DevOther} \\
    \midrule
        {E1} & {None}           & 6.8 & 17.2 & 7.0 &  17.3 \\
    \midrule
        {E2} & {32}             & 6.2 & 14.8 & 6.5 & 15.3 \\
        {E3} & {64}             & 6.5 & 15.1 & 6.8 & 15.5 \\
        {E4} & {128}            & 7.1 & 16.4 & 7.1 & 16.5 \\
    \midrule
        \multicolumn{2}{c|}{\textit{E0 v. E2 \% Rel. WER}}
            & {+29\%} & {+22\%} & {+27\%} & {+26\%} \\
    \bottomrule
    \end{tabular}
    \end{adjustbox}
    \caption{Impact of data-limiting on non-IID training.\label{tab:data_limit}}
\end{table}

Table~\ref{tab:data_limit} illustrates the quality improvement due to data-limiting, bringing WER degradation from over 40\% to less than 30\% relative across all sets.

\subsubsection{Federated Variational Noise}
\label{sssec:federated_vn}

The Baseline used Variational Noise~\cite{variational_inference} (VN), applied by adding Gaussian noise to model parameters during each optimization step. A modification had to be made to VN in order to accommodate the two-step optimization that exists in FL: allowing each client to add its own random noise tensors during local optimization. We called this \textit{Federated} Variational Noise (FVN), and found it was critical to recuperating the non-IID quality degradation. Table~\ref{tab:fvn_convergence} shows experiments \emph{E5} and \emph{E6}, which introduced FVN in a similar manner to the Baseline. We exceeded Baseline model quality by further improving the application of FVN in \emph{E7}, wherein we increased the standard deviation of Gaussian noise linearly during training.

\begin{table}[ht]
    \centering
    \begin{adjustbox}{center, width=\columnwidth-20pt}
    \begin{tabular}{c | c | c c c c}
    \toprule
        \multirow{2}*{\textbf{ID}}
        & \multirow{2}*{\textbf{FVN Std Dev}}
        & \multicolumn{4}{c}{\textbf{WER}} \\
        & & \textit{Test} & \textit{TestOther} & \textit{Dev} & \textit{DevOther} \\
    \midrule
        {E2} &  -               & 6.2 & 14.8 & 6.5 & 15.3 \\
    \midrule
        {E5} & 0.01             & 5.1 & 12.6 & 5.5 & 12.4 \\
        {E6} & 0.02             & 5.0 & 12.2 & 5.2 & 12.4 \\
        {E7} & Ramp to 0.03        & \textbf{4.6} & \textbf{11.9} & \textbf{5.0} & \textbf{11.9} \\
    \midrule
        \multicolumn{2}{c|}{\textit{E0 v. E7 \% Rel. WER}}
            & {-4\%} & {-2\%} & {-2\%} & {-2\%} \\
    \bottomrule
    \end{tabular}
    \end{adjustbox}
    \caption{Impact of FVN on non-IID training.\label{tab:fvn_convergence}}
\end{table}

In addition, we hypothesize that FVN regularizes non-IID client drift because VN was designed to reduce entropy in Bayesian inference tasks. It is based on the idea that model parameters are like random variables sampled from a prior distribution, $\gamma$, which can be better approximated with a given distribution $Q(\beta)$ by adding Gaussian noise during training. Therefore, under FL, if noise from the same underlying Gaussian is applied on each client, the resulting approximation is that all client model parameters are sampled from the same $Q(\beta)$ distribution, thus limiting per-client drift.

\begin{table}[h!]
    \centering
    \begin{adjustbox}{center, width=\columnwidth-20pt}
    \begin{tabular}{c | c | c c c c}
    \toprule
        \multirow{2}*{\textbf{ID}}
        & \multirow{2}*{\textbf{Data Limit}}
        & \multicolumn{4}{c}{\textbf{WER}} \\
        & & \textit{Test} & \textit{TestOther} & \textit{Dev} & \textit{DevOther} \\
    \midrule
        {E7} & 32           & 4.6 & 11.9 & 5.0 & 11.9 \\
    \midrule
        {E8} & -            & 4.6 & 11.9 & 5.1 & 11.8 \\
    \bottomrule
    \end{tabular}
    \end{adjustbox}
    \caption{Impact of data-limiting on FVN experiments.\label{tab:fvn_no_data_limit}}
\end{table}

Results in Table~\ref{tab:data_limit} show that model quality degrades as the per-client data volume is increased. Client models \emph{drift} in varying directions, causing server updates to be sub-optimal. However, experiments \emph{E7} and \emph{E8} in Table~\ref{tab:fvn_no_data_limit} show that, with the addition of FVN, there is minimal change in model quality even without per-client data limits. This adds evidence to our claim that FVN prevents client drift.

\subsection{Computational Efficiency}
\label{ssec:computational_efficiency}

\subsubsection{Quality-Cost Analysis}
\label{sssec:perf_cost_analysis}

So far, we focused on quality impact due to non-IID training. However, as described in Section~\ref{ssec:cost_func}, a crucial aspect of designing an FL system lies in its computational cost. In production FL, model payload size would vary per-experiment due to the presence of transport compression. Likewise, client memory usage would vary across devices due to differing hardware characteristics. Motivated by simplicity, approximations were used in this analysis. The round trip communication payload was approximated to be twice the model size (960 MB), and peak memory was approximated as the size of the model plus 10\% intermediate storage (660 MB). As Eq.~\ref{eqn:fl_cost} also requires $\alpha$ to be set, it was chosen as 1 for this study.

\begin{figure}[htpb]
\begin{minipage}[b]{1.0\linewidth}
  \centering
  \begin{tikzpicture}
    \begin{axis}[
        height=4cm,
		width=8cm,
        legend pos=north east,
        xlabel={Round},
    	ylabel={Mean WER},
    	ytick={11.8, 12, 12.2},
    	ymax=12.2,
    	grid=both,
    	enlargelimits=0.2]
    \draw[dashed, color=blue, line width=0.2mm] (0, 12.1) -- (30000, 12.1);
    \addplot[scatter, color=blue, mark=*,only marks,
            scatter src=explicit symbolic,
            nodes near coords,]
    table[x=ckpt,y=other,meta=id ,col sep=comma]{performance_cost_basic.csv};
    \end{axis}
    \end{tikzpicture}
  \centerline{(a) Comparing experiments by rounds to convergence.}\medskip
\end{minipage}
\hfill
\begin{minipage}[b]{1.0\linewidth}
  \centering
  \begin{tikzpicture}
    \begin{axis}[
        height=4cm,
		width=8cm,
        legend pos=north east,
        xlabel={\textrm{CFMQ} [TeraBytes]},
    	ylabel={Mean WER},
    	ytick={11.8, 12, 12.2},
    	ymax=12.2,
    	grid=both,
    	enlargelimits=0.2]
    \draw[dashed, color=red, line width=0.2mm] (0, 12.1) -- (30000, 12.1);
    \addplot[scatter, color=red, mark=*,only marks,
            scatter src=explicit symbolic,
            nodes near coords,]
    table[x=cost,y=other,meta=id,col sep=comma]{performance_cost_basic.csv};
    \end{axis}
    \end{tikzpicture}
  \centerline{(b) Comparing experiments by cost function.}\medskip
\end{minipage}
\caption{Experiment efficiency comparison.}
\label{fig:cost_scatter}
\end{figure}

Figure~\ref{fig:cost_scatter} shows the quality-cost trade-off for key experiments from the previous section, contrasting number of rounds as the measure of cost against \textrm{CFMQ}. Quality is measured through mean WER on the \emph{Other} evaluation sets, as these are more challenging. If rounds to convergence is the measure of cost, \emph{E8} achieved better quality than the Baseline for a marginal increase in cost. However, when using \textrm{CFMQ}, it is clear that \emph{E7} achieved the same quality as \emph{E8} at lower cost. This is due to the fact that no per-client data limits were imposed in \emph{E8}, leading clients to take more local optimization steps than in \emph{E7} for the same model quality.

\subsubsection{Reducing the Cost of non-IID Model Quality}
\label{sssec:close_cost_gap}

Experiments thus far have recuperated the quality loss due to non-IID training data, but incurred an increase in cost. New experiments, aimed at reducing cost, were conducted by varying the number of local epochs, server learning rate schedule, and amount of SpecAugment. Table~\ref{tab:cfmq} shows the two most promising experiments, \emph{E9} and \emph{E10}, which had a lower \textrm{CFMQ} and better quality in comparison to the Baseline. They both modified the learning rate schedule to have a shorter ramp-up and introduced an exponential decay.~\emph{E10} increased the amount of SpecAugment during the training procedure and yielded slightly better quality. Therefore, we were able to recuperate the quality degradation from non-IID training data at a lower computation cost than the IID Baseline in this study. We must note that in order to limit scope we did not re-visit and refine the Baseline in this work.

\begin{table}[htpb]
    \centering
    \begin{adjustbox}{center, width=\columnwidth-20pt}
    \begin{tabular}{c | c | c c c c}
    \toprule
        \multirow{2}*{\textbf{ID}}
        & \multirow{2}*{\textbf{\textrm{CFMQ} [TB]}}
        & \multicolumn{4}{c}{\textbf{WER}} \\
        & & \textit{Test} & \textit{TestOther} & \textit{Dev} & \textit{DevOther} \\
    \midrule
        {E0}    &   {3077}          & 4.8 & 12.1 & 5.1 &  12.1 \\
    \midrule
        {E9}    &   {2779}          & 4.8 & 11.4 & 4.6 & 11.5 \\
        {E10}   &   {2945}          & 4.8 & 11.4 & 4.6 & 11.4 \\
    \bottomrule
    \end{tabular}
    \end{adjustbox}
    \caption{Exceeding Baseline quality with lower \textrm{CFMQ}.\label{tab:cfmq}}
\end{table}

%% file: conclusions.tex
\section{Conclusion}
\label{sec:conclusion}

Federated learning implies training on non-IID data, a property that has been considered a potential drawback of the technique. We argued that the degree of non-IID-ness can be adjusted through random client data sampling, resulting in a flexible cost-quality trade-off. Initially, recuperating quality in a federated setting is likely to lead to a cost increase. When this is resolved, e.g. through optimizer configuration, hyper-parameter tuning, and use of regularizers, FL can provide IID-level quality at relatively low costs. We demonstrated that this double optimization could be performed for the federated learning of a state-of-the-art ASR model, resulting in a better model at lower cost than the baseline Adam-SGD model.

%% file: acknowledgements.tex
\section{Acknowledgements}
\label{sec:ack}

We would like to thank Khe Chai Sim, Lillian Zhou, Petr Zadrazil, Hang Qi, Harry Zhang, Yuxin Ding, and Tien-Ju Yang for providing valuable insights on its structure and contents.

%% file: paper.bbl
\begin{thebibliography}{10}

\bibitem{device_ownership}
``{Mobile Fact Sheet: Mobile Phone Ownership Over Time},''
  \url{https://www.pewresearch.org/internet/fact-sheet/mobile/}, 2019,
\newblock Accessed: 2020-09-22.

\bibitem{st_rnn}
A.~Graves,
\newblock ``{Sequence Transduction with Recurrent Neural Networks},''
\newblock {\em CoRR}, vol. abs/1211.3711, 2012.

\bibitem{sequence_models_in_asr}
A.~{Graves}, A.~{Mohamed}, and G.~{Hinton},
\newblock ``{Speech Recognition with Deep Recurrent Neural Networks},''
\newblock in {\em 2013 IEEE International Conference on Acoustics, Speech and
  Signal Processing}, 2013, pp. 6645--6649.

\bibitem{e2e_surpasses_server}
T.~N. {Sainath}, Y.~{He}, B.~{Li}, and other,
\newblock ``{A Streaming On-Device End-To-End Model Surpassing Server-Side
  Conventional Model Quality and Latency},''
\newblock 05 2020, pp. 6059--6063.

\bibitem{e2e_streaming}
Y.~{He}, T.~N. {Sainath}, R.~{Prabhavalkar}, et~al.,
\newblock ``{Streaming End-to-end Speech Recognition for Mobile Devices},''
\newblock in {\em ICASSP 2019 - 2019 IEEE International Conference on
  Acoustics, Speech and Signal Processing (ICASSP)}, 2019, pp. 6381--6385.

\bibitem{e2e_asr_blogpost}
Google Inc.,
\newblock ``{An All-Neural On-Device Speech Recognizer},''
  \url{https://ai.googleblog.com/2019/03/an-all-neural-on-device-speech.html},
  2019,
\newblock Accessed: 2020-10-08.

\bibitem{fedavg}
H.~B. McMahan, E.~Moore, et~al.,
\newblock ``{Communication-Efficient Learning of Deep Networks from
  Decentralized Data},''
\newblock in {\em Proceedings of the 20th International Conference on
  Artificial Intelligence and Statistics, {AISTATS} 2017, 20-22 April 2017,
  Fort Lauderdale, FL, {USA}}, A.~Singh and X.~(Jerry) Zhu, Eds. 2017, vol.~54
  of {\em Proceedings of Machine Learning Research}, pp. 1273--1282, {PMLR}.

\bibitem{fl_at_scale}
K.~Bonawitz, H.~Eichner, et~al.,
\newblock ``{Towards Federated Learning at Scale: System Design},''
\newblock in {\em SysML 2019}, 2019,
\newblock To appear.

\bibitem{fedemoji}
F.~Beaufays, K.~Rao, R.~Mathews, et~al.,
\newblock ``{Federated Learning for Emoji Prediction in a Mobile Keyboard},''
  2019.

\bibitem{fed_nextword}
A.~Hard, K.~Rao, R.~Mathews, et~al.,
\newblock ``{Federated Learning for Mobile Keyboard Prediction},''
\newblock {\em CoRR}, vol. abs/1811.03604, 2018.

\bibitem{fed_query}
T.~Yang, G.~Andrew, H.~Eichner, et~al.,
\newblock ``{Applied Federated Learning: Improving Google Keyboard Query
  Suggestions},''
\newblock {\em CoRR}, vol. abs/1812.02903, 2018.

\bibitem{noniid_fedvision}
T.~H. Hsu, H.~Qi, and M.~Brown,
\newblock ``{Measuring the Effects of Non-Identical Data Distribution for
  Federated Visual Classification},''
\newblock {\em CoRR}, vol. abs/1909.06335, 2019.

\bibitem{noniid_keyword}
A.~Hard, K.~Partridge, C.~Nguyen, et~al.,
\newblock ``{Training Keyword Spotting Models on Non-IID Data with Federated
  Learning},'' 2020.

\bibitem{noniid_divergence}
Y.~Zhao, M.~Li, L.~Lai, et~al.,
\newblock ``{Federated Learning with Non-IID Data},''
\newblock {\em CoRR}, vol. abs/1806.00582, 2018.

\bibitem{fl_open_problems}
P.~Kairouz, H.~B. McMahan, et~al.,
\newblock ``{Advances and Open Problems in Federated Learning},''
\newblock 2019.

\bibitem{noniid_quagmire}
K.~Hsieh, A.~Phanishayee, O.~Mutlu, et~al.,
\newblock ``{The Non-IID Data Quagmire of Decentralized Machine Learning},''
  2019.

\bibitem{adaptive_optimizers}
S.~Reddi, Z.~Charles, M.~Zaheer, et~al.,
\newblock ``{Adaptive Federated Optimization},'' 2020.

\bibitem{fed_dga}
D.~Dimitriadis, K.~Kumatani, R.~Gmyr, et~al.,
\newblock ``{Federated Transfer Learning with Dynamic Gradient Aggregation},''
  2020.

\bibitem{fedvision_mitigating_noniid}
T.~H. Hsu, H.~Qi, and M.~Brown,
\newblock ``{Federated Visual Classification with Real-World Data
  Distribution},'' 2020.

\bibitem{libri}
V.~{Panayotov}, G.~{Chen}, D.~{Povey}, and S.~{Khudanpur},
\newblock ``{Librispeech: An ASR Corpus Based on Public Domain Audio Books},''
\newblock in {\em 2015 IEEE International Conference on Acoustics, Speech and
  Signal Processing (ICASSP)}, 2015, pp. 5206--5210.

\bibitem{specaug}
D.~S. Park, W.~Chan, Y.~Zhang, et~al.,
\newblock ``{SpecAugment: A Simple Data Augmentation Method for Automatic
  Speech Recognition},''
\newblock {\em Interspeech 2019}, Sep 2019.

\bibitem{variational_inference}
A.~Graves,
\newblock ``{Practical Variational Inference for Neural Networks},''
\newblock in {\em Advances in Neural Information Processing Systems 24},
  J.~Shawe-Taylor, R.~S. Zemel, P.~L. Bartlett, et~al., Eds., pp. 2348--2356.
  Curran Associates, Inc., 2011.

\bibitem{tensorflow}
M.~Abadi, P.~Barham, , J.~Chen, et~al.,
\newblock ``{TensorFlow: A System for Large-scale Machine Learning},''
\newblock in {\em 12th USENIX Symposium on Operating Systems Design and
  Implementation (OSDI 16)}, 2016, pp. 265--283.

\bibitem{tpu}
N.~P. {Jouppi}, C.~{Young}, et~al.,
\newblock ``{In-datacenter Performance Analysis of a Tensor Processing Unit},''
\newblock in {\em 2017 ACM/IEEE 44th Annual International Symposium on Computer
  Architecture (ISCA)}, 2017, pp. 1--12.

\end{thebibliography}
